\documentclass[10pt,twocolumn,letterpaper]{article}
\usepackage{booktabs}       
\usepackage{multirow}       
\usepackage{threeparttable} 
\usepackage{array} 
\usepackage{graphicx} 
\usepackage{cvpr}              

\definecolor{cvprblue}{rgb}{0.21,0.49,0.74}
\usepackage[pagebackref,breaklinks,urlcolor=cvprblue, linkcolor=cvprblue,colorlinks=true,citecolor=cvprblue]{hyperref}

\def\confName{CVPR}
\def\confYear{2025}

\title{Technical Report for Ego4D Long-Term Action Anticipation Challenge 2025}

\author{
Qiaohui Chu$^{1\,2}$, Haoyu Zhang$^{1\,2}$, Yisen Feng$^{1}$, Meng Liu$^{3}$, Weili Guan$^{1}$, \\ Yaowei Wang$^{1\,2}$, Liqiang Nie$^{1}$\\
$^1$Harbin Institute of Technology (Shenzhen) \qquad  $^2$Pengcheng Laboratory    \\$^3$Shandong Jianzhu University\\
{\tt\small \{qiaohuichu8599, zhang.hy.2019, yisenfeng.hit, mengliu.sdu, honeyguan, nieliqiang\}@gmail.com;} \\ {\tt\small wangyw@pcl.ac.cn}
}

\begin{document}
\maketitle
\begin{abstract}
In this report, we present a novel three-stage framework developed for the Ego4D Long-Term Action Anticipation (LTA) task. 
Inspired by recent advances in foundation models, our method consists of three stages: feature extraction, action recognition, and long-term action anticipation. First, visual features are extracted using a high-performance visual encoder. The features are then fed into a Transformer to predict verbs and nouns, with a verb–noun co-occurrence matrix incorporated to enhance recognition accuracy. Finally, the predicted verb–noun pairs are formatted as textual prompts and input into a fine-tuned large language model (LLM) to anticipate future action sequences.
Our framework achieves first place in this challenge at CVPR 2025, establishing a new state-of-the-art in long-term action prediction.
Our code will be released at \href{https://github.com/CorrineQiu/Ego4D-LTA-Challenge-2025}{https://github.com/CorrineQiu/Ego4D-LTA-Challenge-2025}.
\end{abstract}

\section{Introduction}
\label{sec:intro}

Egocentric video understanding has emerged as a critical research direction, driven by the rapid proliferation of large-scale egocentric video datasets. Unlike traditional third-person videos, egocentric videos capture everyday activities directly from the wearer's perspective, inherently embedding the actor’s attention, actions, and subjective intentions~\cite{zhang2025exo2ego}. This unique viewpoint closely aligns egocentric research with the goals of Embodied AI, where understanding and predicting human behavior from the actor's perspective is essential. However, due to frequent head or body movements and limited environmental context, egocentric videos present distinct and significant analytical challenges, inspiring various complex research problems~\cite{feng2024objectnlq, feng2025object, zhang2023attribute, zhang2024hcqa, zhang2025hcqa, pmlr-v235-zhang24aj}.

A representative task in this area is Long-Term Action Anticipation (LTA) from the Ego4D benchmark\cite{grauman2022ego4d}, which aims to predict an actor’s future behavior from observed video sequences, typically represented as sequences of verb–noun pairs. Compared to third-person LTA tasks, egocentric LTA has greater potential to directly facilitate natural and effective human–computer interaction by predicting future actions from the actor's own viewpoint.

Recently, notable breakthroughs in visual perception and vision–language understanding~\cite{guan2022bi, zhang2021multimodal, zhang2023attribute, zhang2023uncovering, liu2018attentive, liu2018cross}, coupled with advancements in temporal sensitivity of large language models (LLMs)\cite{wang2025time}, have significantly enhanced the models' capabilities in video content understanding and representation. Research in egocentric LTA has gradually shifted from small-scale predictive models to methods based on LLMs. Palm~\cite{huang2023palm} initially demonstrated the feasibility of using Llama2-7B~\cite{touvron2023llama} for action anticipation, while AntGPT improved upon Palm by fine-tuning Llama2-7B, achieving notably better performance. However, as highlighted in both Palm and AntGPT, the accuracy of the initial action recognition models significantly affects the final prediction quality, and existing recognition models still have considerable room for improvement.

To address this issue, we propose an efficient three-stage framework comprising (1) feature extraction, (2) action recognition, and (3) long-term action anticipation. By leveraging a high-performing visual encoder for feature extraction, incorporating hand–object interaction cues, and integrating a verb–noun co-occurrence matrix, our framework improves the accuracy of the action recognition stage. These enhancements collectively strengthen the reliability and precision of the subsequent long-term action anticipation.
Empirical validation demonstrates the effectiveness of our method. Specifically, in the Ego4D Long-Term Action Anticipation Challenge, our approach surpasses all other competitors on the public leaderboard, achieving first place in the challenge.

\begin{figure*}[t]
  \centering
  \includegraphics[width=\linewidth]{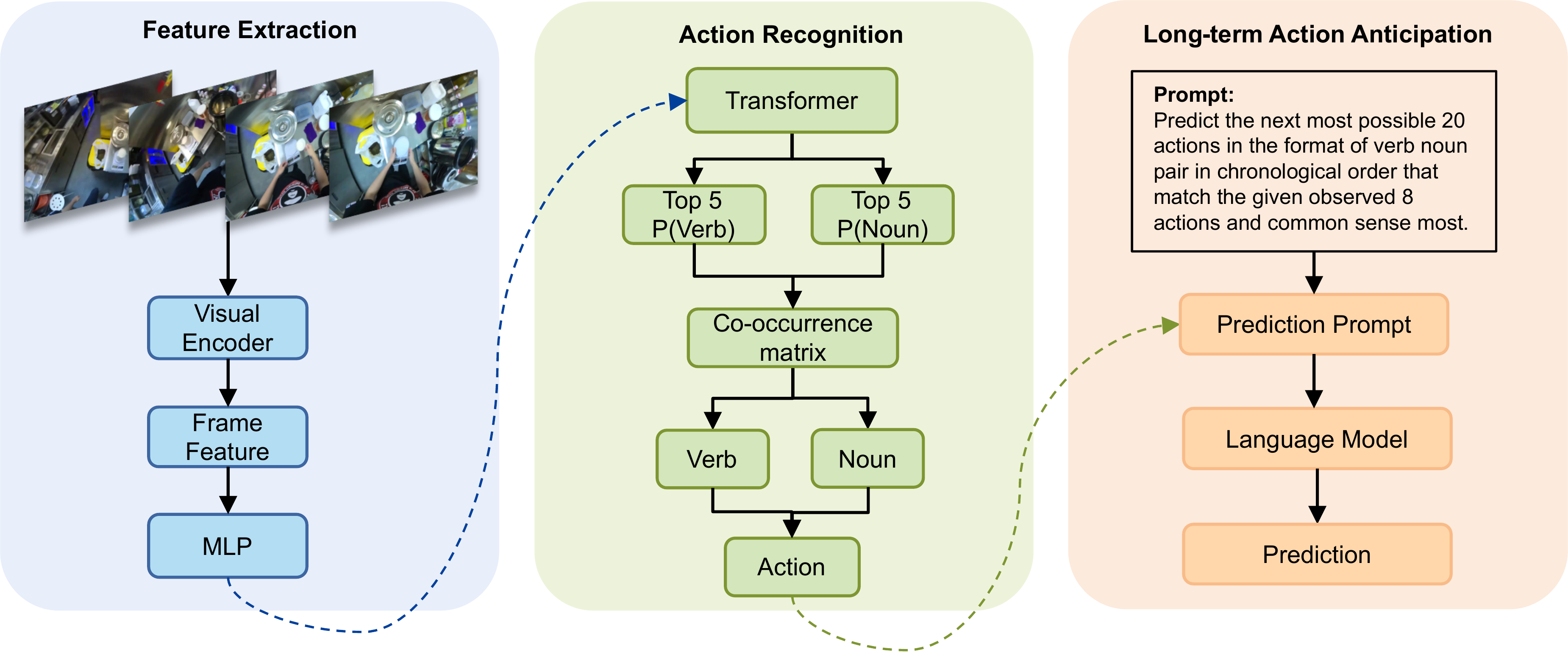}
  \caption{An illustration of our two-stage pipeline.}
  \label{fig:method}
\end{figure*}

\section{Methodology}
\label{sec:formatting}

Our proposed method, as illustrated in the Figure ~\ref{fig:method},
is composed of three main stages: feature extraction, action recognition, and long-term action anticipation.

\subsection{Feature Extraction}
In the feature extraction stage, we adopt the general framework introduced by AntGPT~\cite{zhao2023antgpt}, uniformly sampling four frames from each video segment that corresponds to an action. After evaluating two visual encoders, EgoVLP~\cite{lin2022egocentric} and EgoVideo-V~\cite{pei2024egovideo}, we select EgoVideo-V due to its superior performance for extracting visual features. To enhance the quality of input features, we incorporate a segmentation-based method~\cite{ravi2024sam} to extract visual information from hand–object interaction regions. Both the hand–object interaction regions and the original video frames are fed into the visual encoder to obtain their respective features, which are then fused using a lightweight MLP. This enriched input helps improve the downstream action recognition accuracy.

\subsection{Action Recognition}

In the action recognition stage, visual features are fed into a Transformer model to predict candidate verbs and nouns, from which we obtain the top five predictions for both verbs and nouns. Inspired by the verb-noun co-occurrence matrix proposed in QueryMamba~\cite{zhong2024querymamba}, we further integrate a similar co-occurrence matrix into our recognition model to enhance prediction accuracy. Specifically, we multiply the probability distributions of the top-five verbs and nouns output from the Transformer model by a normalized co-occurrence matrix. Each top-five verb generates five noun candidates with the highest co-occurrence probabilities, forming the top-five verb–noun pairs. Similarly, each top-five noun generates five verb candidates, forming another set of top-five noun–verb pairs. From these ten candidate verb–noun pairs, we select the pair with the highest overall probability as the final prediction of the action recognition stage.

\subsection{Long-term Action Anticipation}

In the action anticipation stage, existing studies typically convert visual observation histories into textual form using an action recognition model, and then utilize LLMs for inference, as demonstrated by methods such as Palm, AntGPT, and EgoVideo. Notably, AntGPT and EgoVideo achieved improved results by fine-tuning Llama2-7B and Vicuna-7B~\cite{zheng2023judging}, respectively. Following a similar approach, we feed the textual representations of visual observation histories obtained from the action recognition stage into a fine-tuned Llama2-7B model, predicting a sequence of 20 future actions. 

\begin{table*}[htbp]
  \centering
  \caption{Performance comparison of existing work and the top six teams on the public leaderboard. The best results are shown in bold.}
  \label{tab:leaderboard}
  \begin{threeparttable}
  \begin{tabular*}{\textwidth}{@{\extracolsep{\fill}}
      c c c c c @{}}
    \toprule
    \multirow{2}{*}{Method} 
      & \multirow{2}{*}{Rank} 
      & \multicolumn{3}{c}{LTA ED$\downarrow$ } \\
    \cmidrule(lr){3-5}
    & 
      & Verb
      & Noun
      & Action \\
    \midrule
    VideoLLM~\cite{chen2023videollm}       & -- & 0.7210 & 0.7250 & 0.9210 \\
    Palm (Challenge)~\cite{huang2023palm} & -- & 0.6956 & 0.6506 & 0.8856 \\
    AntGPT~\cite{zhao2023antgpt}         & -- & 0.6503 & 0.6498 & 0.8770 \\
    PALM (Paper)~\cite{kim2024palm}   & -- & 0.6471 & 0.6117 & 0.8503 \\
    Charlie’s (AntGPT) & 6 & 0.6531 & 0.6446 & 0.8748 \\
    HAI-PUI        & 5  & 0.6362 & 0.6258 & 0.8663 \\
    Doggeee (w/o finetuning) & 4 & 0.6663 & 0.6240 & 0.8650 \\
    BigMac (mtp)   & 3  & \textbf{0.6340} & 0.6395 & 0.8649 \\
    123456ABCD (newdict)(EgoVideo)~\cite{pei2024egovideo} & 2 & 0.6354 & 0.6367 & 0.8504 \\
    Ours (iLearn2.O) & 1 & 0.6346 & \textbf{0.5986} & \textbf{0.8493} \\
    \bottomrule
  \end{tabular*}
  \end{threeparttable}
\end{table*}

\section{Experiment}

\subsection{Performance Comparison}
The performance on the LTA task is evaluated using the Edit Distance (ED) metric, which measures the average number of insertions, deletions, substitutions, and transpositions between the predicted sequences and the ground truth sequences. Table~\ref{tab:leaderboard} summarizes results from existing LLM-based methods and the public leaderboard on the Ego4D v2 test set. As shown in the table, our proposed framework achieves the best (ranked first) ED scores for both nouns and overall actions. In terms of verb ED, our method ranks second, with only a slight difference of 0.06\% compared to the top-ranked method. These results clearly demonstrate the superior performance of our framework compared to existing approaches.

\begin{table*}[htbp]
  \centering
  \begin{threeparttable}
    \caption{Performance comparison of our method and baselines. † indicates using a sliding-window strategy; ‡ indicates the model is finetuned on the Ego4D dataset. In the AR table, bolded results indicate performance using a zero-shot visual encoder without applying the sliding window strategy. In the LTA table, bolded results represent the best-performing scores.}
    \label{tab:performance}
    \begin{tabular*}{\textwidth}{@{\extracolsep{\fill}}
        c  c  c  c  c  c  c  c  c @{}}
      \toprule
      \multirow{2}{*}{Method}
        & \multirow{2}{*}{LLM}
        & \multirow{2}{*}{Visual Encoder}
        & \multicolumn{3}{c}{AR Acc\,$\uparrow$}
        & \multicolumn{3}{c}{LTA ED\,$\downarrow$} \\
      \cmidrule(lr){4-6}\cmidrule(lr){7-9}
      & & 
        & Verb  & Noun   & Action
        & Verb
        & Noun
        & Action\\
      \midrule
      \multirow{2}{*}{PALM~\cite{kim2024palm}} 
        & \multirow{2}{*}{LLaMA2-7B~\cite{touvron2023llama}}
        & EgoVLP~\cite{lin2022egocentric}
        & 32.87 & 38.26  & 15.18
        & --    & --     & --     \\
        &
        & EgoVLP\tnote{\dag}  
        & 40.32 & 45.53  & 20.63
        & 0.6471 & 0.6117 & 0.8503 \\

      Egovideo~\cite{pei2024egovideo}
        & Vicuna-7B~\cite{zheng2023judging}
        & EgoVideo-V\tnote{\ddag}
        & 43.65 & 52.21  & 27.64
        & 0.6354 & 0.6367 & 0.8504 \\

      Ours (EgoVLP)
        & LLaMA2-7B
        & EgoVLP
        & 30.90 & 40.28  & 15.36
        & 0.6491 & 0.6429 & 0.8718 \\

      Ours (EgoVideo-V)
        & LLaMA2-7B
        & EgoVideo-V
        & \textbf{38.45} & \textbf{51.50}  & \textbf{22.15}
        & \textbf{0.6346} & \textbf{0.5986} & \textbf{0.8493} \\
      \bottomrule     
    \end{tabular*}
  \end{threeparttable}
\end{table*}

\subsection{Ablation Study}

Table~\ref{tab:performance} shows ablation results for different visual encoders and inference LLMs. The results show that our approach consistently outperforms baseline methods in recognition accuracy, regardless of whether EgoVLP or EgoVideo-V is used as the backbone encoder. Furthermore, we observe that the advanced EgoVideo-V achieves the highest action recognition accuracy, reaching 22.15\%.

For the long-term action anticipation, we utilize a fine-tuned Llama2-7B as our anticipation model. Despite the action recognition accuracy achieved by our model being slightly lower than the highest accuracy from the fine-tuned EgoVideo-V, our overall LTA predictions clearly outperform all baselines. By combining improvements across all three stages, we achieve superior overall performance on the LTA task, demonstrating the positive contributions of each component to the final prediction accuracy.

\begin{figure}[htbp]
  \centering
  \includegraphics[width=\linewidth]{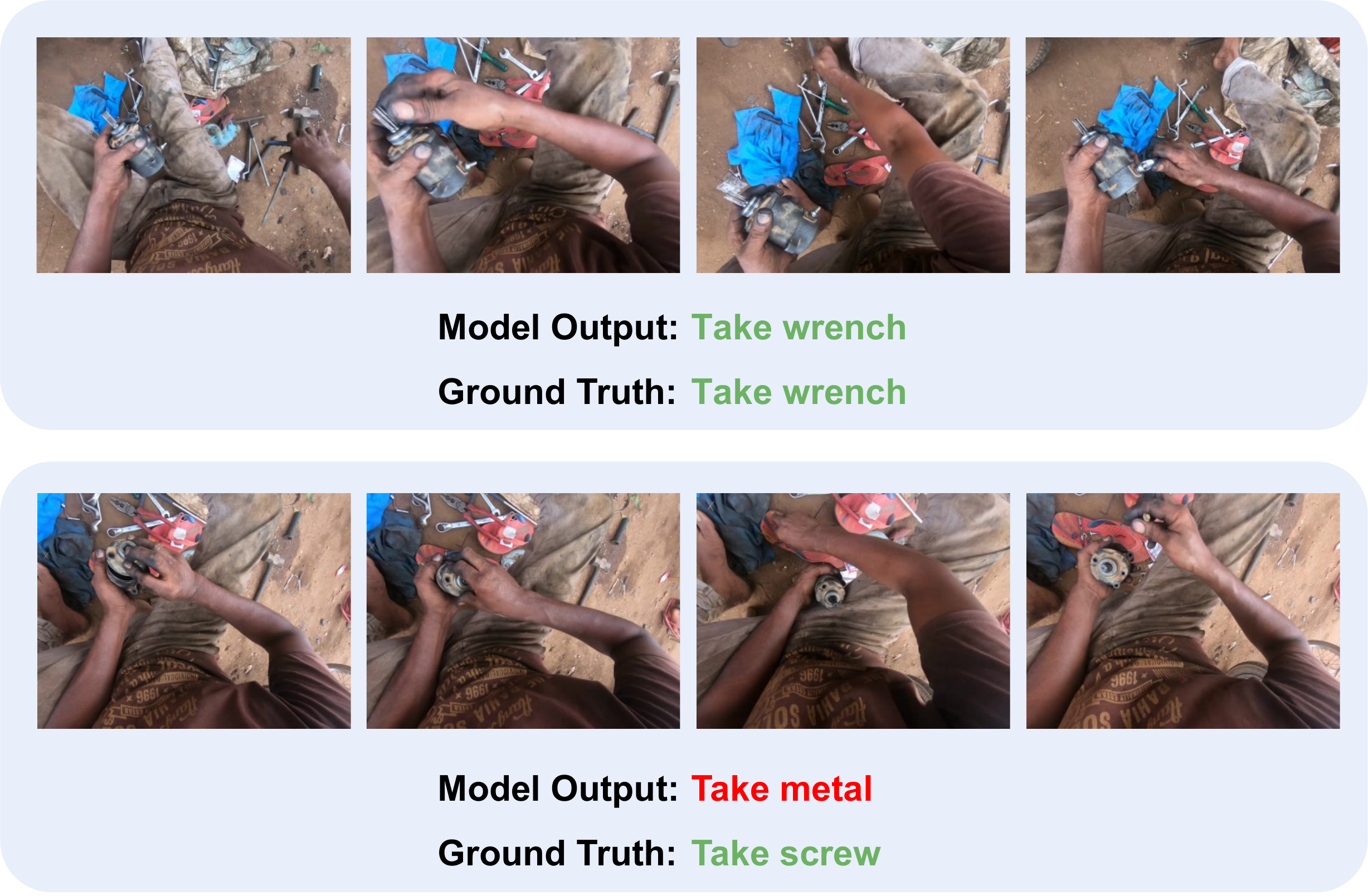}
  \caption{Successful and failed cases of Action Recognition model.}
  \label{fig:AR_case}
\end{figure}

\begin{figure}[htbp]
  \centering
  \includegraphics[width=\linewidth]{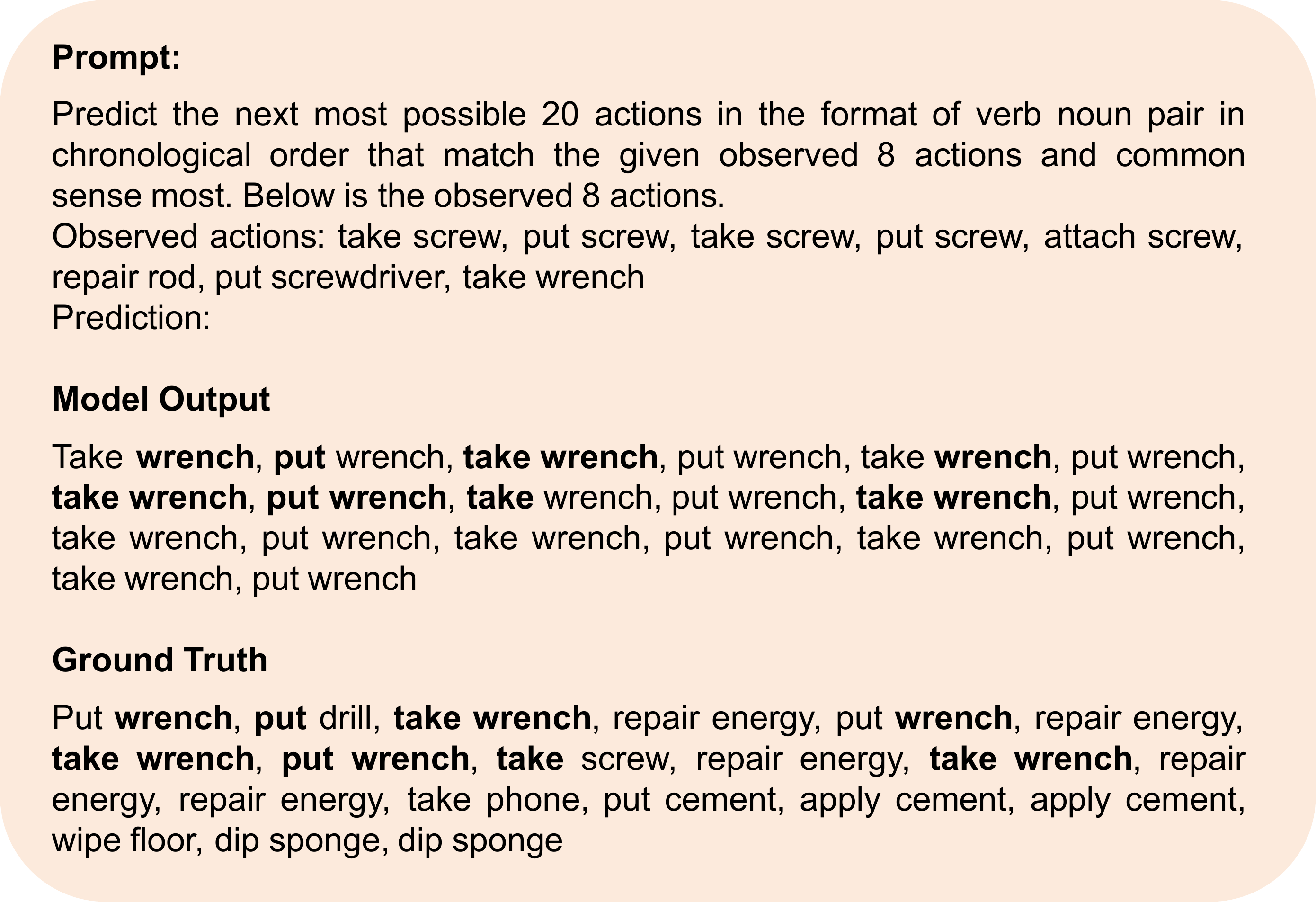}
  \caption{An example from our action anticipation model.}
  \label{fig:LTA_case}
\end{figure}

\subsection{Case Study}
Figure~\ref{fig:AR_case} illustrates both successful and failed examples from our action recognition model. In the failed action recognition case, we notice the model struggles to differentiate between semantically related nouns or verbs with varying levels of granularity. For example, the model incorrectly identifies “screw” as “metal”; although a screw is indeed a type of metal, the difference in granularity leads to incorrect predictions. Addressing this limitation in future work may involve methods to enhance the model’s ability to recognize actions at different semantic granularities.

Figure~\ref{fig:LTA_case} illustrates an example from our action anticipation model. For this failed example, the model demonstrates primarily insufficient predictive capability for future scenarios, resulting in repetitive sequences closely related to the final observed actions. For instance, future actions could involve “take phone,” “put cement,” and “apply cement,” but our model failed to anticipate these tasks, leading to repetitive or incorrect predictions. To mitigate this, we plan to incorporate task-intent recognition modules in future work, providing the model with clearer intent-driven guidance for more accurate predictions.

\section{Conclusion}
We introduce a three-stage enhanced framework for the LTA task, which substantially improves performance in both the action recognition and long-term action anticipation. Experimental results demonstrate that our method achieves outstanding performance, securing first place in the CVPR 2025 Ego4D Long-Term Action Anticipation Challenge and surpassing previous state-of-the-art approaches.
{
    \small
    \bibliographystyle{ieeenat_fullname}
    \bibliography{main}
}


\end{document}